\documentclass{article}

 \usepackage[preprint]{neurips_2019}

\usepackage[preprint]{neurips_2019}
\usepackage[T1]{fontenc}    
\usepackage{hyperref}       
\usepackage{url}            
\usepackage{booktabs}       
\usepackage{amsfonts}       
\usepackage{nicefrac}       
\usepackage{microtype}      
\usepackage{float}
\usepackage{array}
\usepackage[utf8]{inputenc}
\usepackage{graphicx}
\usepackage{amsmath}
\usepackage{amssymb}
\usepackage{booktabs,multirow}
\usepackage{listings}
\title{DeepSumm - Deep Code Summaries using Neural Transformer Architecture}

\author{
  Vivek Gupta\\
  \texttt{vkgupta@stanford.edu} \\
}

\begin{document}
\maketitle
\begin{abstract}
  Source code summarizing is a task of writing short, natural language descriptions of source code behaviour during run time. Such summaries are extremely useful for software development and maintenance but are expensive to manually author, hence it is done for small fraction of the code that is produced and is often ignored. Automatic code documentation can possibly solve this at a low cost. This is thus an emerging research field with further applications to program comprehension, and software maintenance. Traditional methods often relied on cognitive models that were built in the form of templates and by heuristics and had varying degree of adoption by the developer community. But with recent advancements, end to end data-driven approaches based on neural techniques have largely overtaken the traditional techniques. Much of the current landscape employs neural translation based architectures with recurrence and attention which is resource and time intensive training procedure. In this paper, we employ neural techiques to solve the task of source code summarizing and specifically compare NMT based techniques to more simplified and appealing Transformer architecture on a dataset of Java methods and comments. We bring forth an argument to dispense the need of recurrence in the training procedure.  To the best of our knowledge, transformer based models have not been used for the task before. With supervised samples of more than 2.1m  comments and code, we reduce the training time by \textbf{more than 50\%} and achieve the BLEU score of \textbf{17.99} for the test set of examples. 
  
\end{abstract}

\section{Introduction}
A “summary” of source code is a brief natural language description of that section of source code. One of the most common targets for summarization are the subroutines or more generally known methods in a program; for example, the one-sentence descriptions of Java methods widely used in automatically-formatted documentation e.g. JavaDocs as shown below:

\textbf{Method}:
\begin{lstlisting}
    public void render(GameData data) 
    {
        setText(Message.render(data, type.getPattern(), attributes))
    }
\end{lstlisting}

\textbf{Comment}:
\begin{verbatim}
    Renders the message and updates the message text
\end{verbatim}
These comments are useful because they help programmers understand the role that the method plays in a program. Empirical studies have repeatedly shown that the  understanding the role of the methods in a program is a crucial step to understanding the program’s behavior overall. Even a short summary of a method such as shown above can tell a programmer a lot about that method and the program as a whole.

A holy grail of software engineering research has long been to generate these summaries automatically. Forward \textit{et al.}. \cite{forward} pointed out in 2002 that “software professionals value technologies that improve automation of the documentation process,” and “that documentation tools should seek to better extract knowledge from core resources”, such as source code [7]. Tools such as Resharper, JavaDoc and Doxygen automate the format and presentation of documentation, but still leave programmers with the most labor-intensive effort of writing the text and examples.

The research task of generating such summaries is also known as "source code summarization" with much emphasis on summarization of methods. For several years, significant progress was made based on content selection and sentence templates but all these techniques have lately given the way to AI based systems based on big data input.

The inspiration for a vast majority of efforts into AI-based code summarization originates in neural machine translation (NMT) from the natural language processing research community. It is typically thought of in terms of sequence to sequence(seq2seq) learning. In software engineering research, machine translation can be considered as a metaphor for source code summarization: the words and tokens in the body of a method are one sequence, while the desired natural language summary is the target sequence. This application of NMT to code summarization has shown strong benefits in a variety of applications. Much of NMT based methods used for summarization task uses attention based architecture which is based on the method to jointly align and translate - which is either based on recurrence or on convolutions. These underlying architectures have systematic problems in them such as exploding or vanishing gradient problem to address long range dependencies. They are computational complex and are not very interpretable. With this project we aim to solve such problem using Transformer based architecture and compare its performance with traditional recurrent, encoder-decoder based seq2seq model with attention.

\section{Related Work}

\textit{{Early methods - Heuristic based/ Template based methods}}

Hauic \textit{et al.} \cite{HaiducAMM10} are often credited with the first attempt to create textual summaries of the code (class or a method) and were the first one to coin the term 'source code summarization'. This early work applied TR methods like Latent Semantic Indexing and combined them with structural information of the code to make effective summaries of the code. A different work was developed by Sridhar \textit{et al.} \cite{Moreno2013AutomaticGO} presented a new method to automatically produce descriptive summary comments for software methods. Based on the method’s signature and body, their comment generator found the content of the summary and produced the natural language text that summarizes the method. Moreno \textit{et al.}.\cite{Moreno2013AutomaticGO} suggested a new method to automatically produce natural language summaries for software classes by using stereotype of class.

As in other research areas related to natural language generation, data-driven techniques have largely supplanted template-based techniques due to a much higher
degree of flexibility and reduced human effort in template creation.

\textit{{End-to-end data driven methods}}

Iyer \textit{et al.} \cite{iyer-etal-2016-summarizing} presented an end to end natural language generation system called CODE-NN that jointly performs content selection using an attention mechanism, and surface realization using Long Short Term
Memory (LSTM) networks. The system generates a summary one word at a time, guided by an attention mechanism over embeddings of the source code, and by context from previously generated words provided by a LSTM network. The simplicity of the
model allows it to be learned from the training data without the burden of feature engineering (Angeli \textit{et al.}., 2010 \cite{angeli-etal-2010-simple}) or the use of an expensive approximate decoding algorithm \cite{Ioannis}. Much of our work is based on the initial setup done by Iyer \textit{et al.} \cite{iyer-etal-2016-summarizing} but incorporating recent advancements in achieving the state of the art models. Of note is that the attentional encoder-decoder seq2seq model originally described by Bahdanau \textit{et al.}. \cite{bahdanau2014neural} is at the core of many of these papers, as it provides strong baseline performance even for many software engineering tasks.

Wei \textit{et al.} \cite{wei2019code} explored code summarization task along with code generation task holistically to produce a model that performs summarization task. They exploit the duality between these tow tasks and a propose a dual training framework to train the tow tasks simultaneously.  They consider the dualities on probablity and attention weights, and design corresponding regularization terms to 
constrain the duality.

Hu \textit{et al.} \cite{Hu} explored the use of AST - Abstract Syntax Trees which
capture structures and semantics of Java methods. ASTs are converted into sequences before they are fed into atypical encoder-decoder based seq2seq model. The model
itself is an off-the-shelf encoder-decoder; the main advancement is the AST-annotated representation called Structurebased Traversal (SBT). SBT is essentially a technique for flattening the AST and ensuring that words in the code are associated with their AST node type. Alex \textit{et al.}\cite{alex2019neural} further exploited this concept and combined this with an attentional encoder-decoder
system, except with two encoders: one for code/text data and one for AST data. Alex \textit{et al. } \cite{alex2019recommendations} also released the dataset which we use in our work.

\section{Approach}

In this work we mainly consider two set of model architectures. The first architecture, the Neural Machine Translation based encoder-decoder (seq2seq) architecture with attention and the secondly Transformer architecture. Seq2seq based models provides baseline results to compare and contrast with Transformer model. 

\subsection{Neural Machine Translation Architecture} \label{nmt}

The workhorse of most Neural Machine Translation (NMT) systems is the attentional encoder-decoder architecture \cite{luong2015effective} . This architecture originated in work by Bahdanau \textit{et al.} \cite{bahdanau2014neural} and is explained in great detail by a plethora of very highly regarded sources such as \cite{luong2015effective}, \cite{sutskever2014sequence}. In this section, we cover only the concepts necessary to understand our approach at a high level.

In an encoder-decoder architecture, there are a minimum of
two recurrent neural networks (RNNs). The first, called the encoder, converts an arbitrary-length sequence into a single vector representation of a specified length. The second, called the decoder, converts the vector representation given by the
encoder into another arbitrary-length sequence. Encoder generally as a bidirectional RNN while decoder is a unidirectional RNN. The sequence inputted to the encoder is one language e.g. English, and the sequence from the decoder is another language e.g. French. For our experiments, we modeled functions as the source sequence and comments as the target sequence to be fed to this recurrent and attention based encoder-decoder architecture. The final hidden states of the encoder are concatenated and set as the initial state of the decoder after a linear projection. With the encoder hidden states at each step and the decoders hidden state at each step we compute a multiplicative attention score. We compute a softmax over these scores and multiply the scores with the hidden states and accumulate them to form the attention score. Attention score is concatenated with the decoder hidden states and sent through a linear layer and we compute the softmax over the possible words of the output vocabulary to produce the next word prediction. 
\begin{figure}[htp]
    \centering
    \includegraphics[width=10cm]{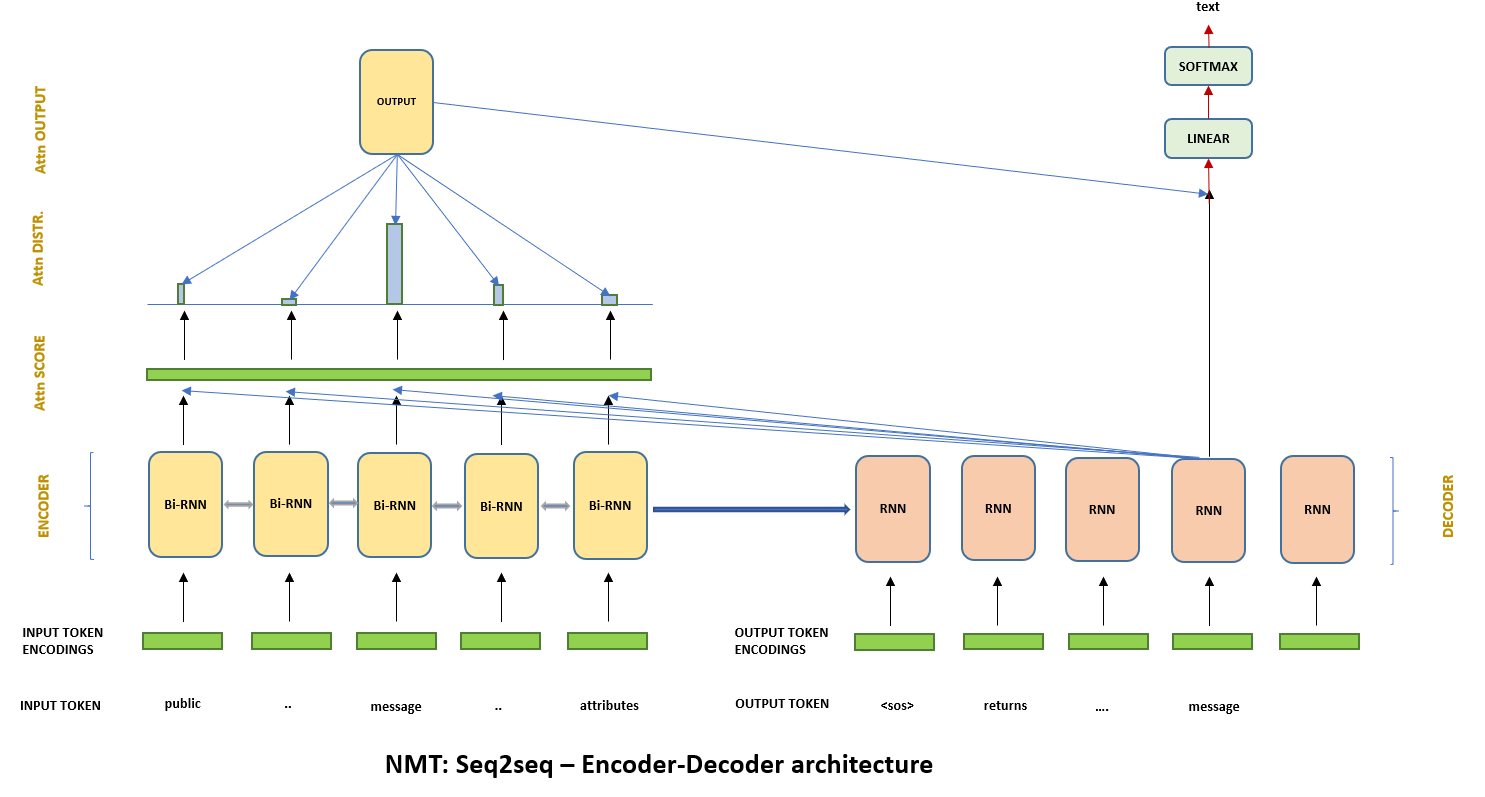}
    \caption{NMT:Seq2seq architecture}
    \label{fig:Network}
\end{figure}

\subsection{Our approach - Transformer architecture} \label{tf}

Much of our work is based on the seminal work done by Vaswani \textit{et al. } \cite{vaswani2017attention}. It is built by many multi-head self attentions blocks. We will delve into the details of this architecture in this section.
\subsubsection{Encoder and Decoder Stacks}
The encoder is composed of a stack of N identical layer. N is a parameter that can be tuned. Each stack of encoder has a multi-head attention layer followed by position wise feed forward network. A residual connection is applied follwoed by layer normalization \cite{ba2016layer} as shown in the figure \ref{fig:Transformer}

Similarly, the decoder is composed of N identical layers and has same set of layers as an encoder but in between the two layers is inserted a masked multihead attention layer that attends to the output of the final encoder layer. 

\subsubsection{Attention functions}
An attention function can be described as mapping a query and a set of key-value pairs to an output, where the query, keys, values, and output are all vectors. The output is computed as a weighted sum of the values. In scaled dot product attentions, the  input consists of queries and keys of dimension \textit{$d_k$}, and the values of dimension \textit{$d_v$}. We compute dot products of the query with all the keys, scale down the results by a factor of $\sqrt{\textit{$d_k$}}$, and apply a softmax to obtain the weights on the values. While implementing it in vectorized form, we compute this by packing all the queries in a matrix, \textit{Q}. The keys are also packed in the matrices \textit{K} and \textit{V}. Since this is computed on the same sequence, we call this function as a Self Attention. Thus, 

\begin{equation}
  \text{Self-Attention(\textit{Q,K,V})  = softmax($\frac{QK^T}{\sqrt{d_k}}$)}V
\end{equation}

Instead of performing a single self attention function, Vaswani \textit{et al.} \cite{vaswani2017attention} found beneficial to to have multiple instances of these self attention functions that operate on multiple (\textit{h} times) projections of the \textit{Q, K} and \textit{V}. The outputs are then concatenated and once again projected, resulting in the final values. 
\begin{equation}
  \text{MultiHead Attention(\textit{Q,K,V})  = Concat(head$_1$,\dots, head$_h$)}W^O
\end{equation}
where each head$_i$  
\begin{equation}
  \text{head$_i$ =  Self-Attention(\textit{QW$_i{^Q}$,KW$_i{^K}$,VW$_i{^V}$})}
\end{equation}
Where the projections are parameter matrices 
\begin{itemize}
    \item W$_i{^Q \in \mathbb{R}^{d_{model}\times{d_{k}}}}$
    \item W$_i{^K \in \mathbb{R}^{d_{model}\times{d_{k}}}}$
    \item W$_i{^V \in \mathbb{R}^{d_{model}\times{d_{v}}}}$
    \item W${^O \in \mathbb{R}^{hd_v\times{d_{model}}}}$
\end{itemize}

Using our experiments, we tune for \textit{h}, \textit{$d_k$} , \textit{$d_v$} , \textit{$d_{model}$} 

The overall flow of tensors is shown in the figure \ref{fig:Transformer}:
\begin{figure}[htp]
    \centering
    \includegraphics[width=15cm]{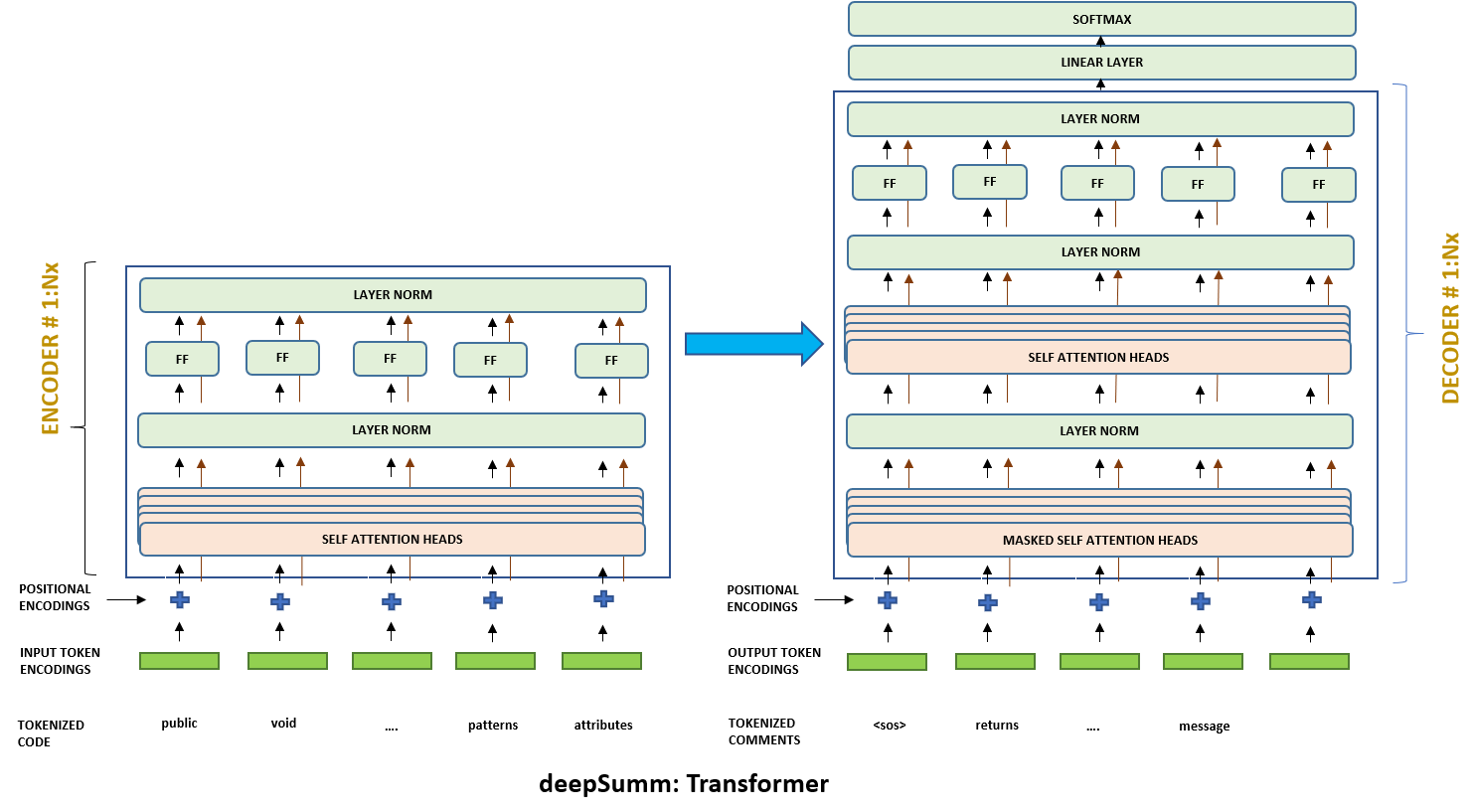}
    \caption{Transformer: Flow of tensors}
    \label{fig:Transformer}
\end{figure}

\subsection{Threats to the approach}
We acknowledge the following issues or threats to the approach: 
\begin{itemize}
    \item The results are very much biased to the Java dataset that is used to train the model. With different dataset, the results could be different.
    \item We were not able to exploit structural composition of the methods and solely relied on the resulting tokens and ignored casing. In production scenario, both these are very important considerations to the comments that go with the code. Our main objective, however is to compare recurrent versus transformer based sequence trasduction procedure.
\end{itemize}

\section{Experiments}

\subsection{Data}
The dataset we use in this paper is based on the dataset provided by LeClair \textit{et al.} \cite{alex2019neural}. We used this dataset because it is both the largest and most recent in source code summarization. LeClair \textit{et al.} provided the dataset after minimal initial processing that filtered for Java methods with JavaDoc comments in English, and removed methods over 100 words long and comments >13 and <3 words. The result is a dataset of 2.1m Java methods and associated comments. Furthermore, LeClair \textit{et al.} provided 3 groups of data, 
\begin{itemize}
    \item Raw data set: 51m Java methods along with original source files.
    \item Filtered data set: 2.1m Java method and comments with unprocessed source code and unprocessed comments.
    \item Tokenized data set: 2.1m Java method and comments. Preprocessed source code with special characters removed, camel case split, lowercased. Comments are the first line of the javadoc lowercased with special characters removed. Our Train, validation and test sets were formed from this group of the data. 
\end{itemize}
Due to the limited resources for the work, we did not use the entire dataset but only a partial set of it for the large runs of experiments. We randomly selected three training sets from the original dataset which are described in the below table and ran various experiments on them to test and validate both the approaches.

\begin{table}[H]
  \caption{Supervised Data set}
  \label{dataclass}
  \centering
  \begin{tabular}{llll}
    \toprule
    \cmidrule(r){1-4}
    Set Label     & Training     & Validation     & Testing \\
    \midrule
    Small Set   & 100,000          & 3000          &3000 \\
    Medium Set  & 1m                & 5000          &5000      \\
    Large Set   & 2.1m       & 10,000       & 10,000 \\ \bottomrule
    
  \end{tabular}
\end{table}

We target the problem of source code summarization of methods – automatic generation of natural language descriptions of methods. Specifically, we target summarization of Java methods, with the objective of creating method summaries like those used in JavaDocs. While we limit the scope of the experiments in this paper to Java dataset, in principle the techniques described in this paper are applicable to any programming language. 

\subsection{Evaluation method}

To evaluate the performance of the model in train and validation sets, we use perplexity measure and to evaluate a model in the test set we use BLEU score \cite{Bleu}.

Perplexity is the measure of how well a probability distribution or a model such as ours, in a train and validation setting, predicts a sample. It is used to compare probabilistic natural language models . a low perplexity indicates the probability distribution is good at predicting the sample. The benefit of using perplexity is that it is easier to calculate. 

To measure the performance of the model we used BLEU score or BiLingual Evaluation Understudy \cite{Bleu}. BLEU is a measure of the number of matching Ngrams between the machine’s output and a reference translation. BLEU has been the traditional standard for MT systems and is considered to be the correspondence between a machine’s output and that of a human. Technically speaking we used Pytorch's, Torchtext implementation of BLEU \cite{bleu_score}, \verb|torchtext.data.metrics.bleu_score|

\subsection{Experimental details}
To get results greedily, we performed a screening test on the hyperparameters and model configurations. To accomplish this we used Small dataset in the table \ref{dataclass} and ran many experiments to get most likelihood set of hyper parameters that might work better. 
We further screened the parameters obtained in the Medium dataset in table \ref{dataclass}. We finally consumed these hyperparameters and configurations in the model with Large dataset.

We used grid search on learning rate, encoder, decoder layer(s), encoder, decoder heads and dimensionality of the model. For the final set of results, we fixed learning rate to the best and the stable rate from the experiments we conducted. 

The baselines used for the model was from the work done by LeClair \textit{et al.} \cite{alex2019neural} in producing models namely \textbf{ast-attendgru} and \textbf{attendgru}. LeClair \textit{et al.} exploited the structural composition of the tokens in the methods along with the tokens themselves in ast-attendgru model, and for attendgru they used a vanilla off the shelf attention based seq2seq model. Since we wanted to experience the difference between recurrent based and a transformer based architecture, deepsumm-attention from section \ref{nmt} on small dataset was our another baseline. deepsumm-transformer from section \ref{tf} and figure \ref{fig:Transformer} was the model under observation for the 3 different classes of our dataset. 

\subsection{Results}

\subsubsection{Baseline}
\begin{table}[h]
\caption{Baselines}
\label{Resultsummary}
    \centering
    \begin{tabular}{ll r }
     \toprule Model & Dataset & BLEU  \\\midrule
            ast-attendgru & Large & 19.6 \\
         attendgru & Large & 19.4 \\
         deepsumm-attention & Small & 10.70 \\ \addlinespace \midrule \addlinespace
         
         deepsumm-transformer & Small & 11.86 \\
         deepsumm-transformer & Medium & 16.95 \\
         deepsumm-transformer & Large & 17.99 \\\bottomrule
         
    \end{tabular}
\end{table}

\subsubsection{Detailed results}
\textit{\textbf{N}}: Encoder, Decoder layers, 
\textit{\textbf{$d_{model}$}}: Input dimension of the token embedding,
\textit{\textbf{batch}}: Batch size, 
\textit{\textbf{h}}: encoder, decoder heads, 
\textit{\textbf{Size}}: Number of parameters(x\textit{$10^6$}), 

\begin{table}[h]
\label{Main}
\caption{deepSumm - Transformer}
  \centering
  \begin{tabular}{lrrrrrrrrrrrr}
    \toprule
    \textbf{\textit{Dataset}} & \textit{\textbf{N}} &\textit{\textbf{$d_{model}$}}&\textit{\textbf{batch}}&\textit{\textbf{h}}&\textit{\textbf{$d_k$,$d_v$}}&\textit{\textbf{Epochs}}&\textit{\textbf{$PPL_{test}$}}&\textit{\textbf{BLEU}}&\textit{\textbf{Size}} \\ \midrule
    \multirow{2}{*}{\textit{\textbf{Small}}} &3&256&128&8&512&10&33.102&11.43&17.5      \\
                          &2&&&&&&30.937&11.8&16.3       \\ \addlinespace
                          &4&&&&&&43.732&9.45&\textit{\textbf{18.7}}       \\ \addlinespace
                          &1&&&&&&32.563&\textit{\textbf{11.86}}&14.8       \\ \addlinespace
                          &2&128&&&&&35.088&11.63&7.7       \\ \addlinespace
                          &1&&&&&&35.854&11.35&7.2       \\ \addlinespace
                          &1&512&&&&&30.47&11.84&31.1       \\ \addlinespace
                          &2&&&&&&\textit{\textbf{30.42}}&11.59&35.3       \\ \addlinespace
                          &2&&&4&&&31.186&11.78&16.3       \\ \addlinespace \bottomrule \addlinespace
    \multirow{2}{*}{\textit{\textbf{Medium}}} &3&256&128&8&512&5&18.003&16.41&\textit{\textbf{53.5}}      \\\addlinespace                                     &2&&&&&&\textit{\textbf{17.928}}&16.8&52.2      \\\addlinespace
                         &1&&&&&&19.415&16.32&50.9      \\\addlinespace  
                         &2&&&4&&&18.376&\textit{\textbf{16.95}}&52.2       \\ \addlinespace  \bottomrule \addlinespace
    \multirow{2}{*}{\textit{\textbf{Large}}} &3&256&256&8&512&5&\textit{\textbf{16.152}}&17.91&\textit{\textbf{76.6}}      \\\addlinespace                                     &2&&&&&&16.431&17.81&75.3      \\\addlinespace
                         &1&&&&&&17.419&17.81&75.3      \\\addlinespace  
                         &2&&&4&&&16.539&\textit{\textbf{17.99}}&75.2      
                         
    \\ \bottomrule
  \end{tabular}
\end{table}

We expected and observed a drop in training time of the deepsumm transformer based models from its attention based recurrent architectures by atleast half. This enabled us to run the experiment on the larger dataset. This is inline with what we learnt from the work done by Vaswani \textit{et al.} \cite{vaswani2017attention}. 
We also hoped to deepsumm transformer model to outperform the recurrent based baseline in both processing time and the performance metric. The results obtained were in line. 

We were however not able to come close to the work done and results obtained by LeClair on the same dataset \cite{alex2019neural}. We suspect not having structural composition embedded into the model is contributing to the loss. We also strongly suspect that our training time and exploration of the gradient space via a static learning rate is not optimal and may also have come strongly contributed to the shortcoming. We were only able to train four models for the larger dataset and for only 5 epochs. 

\section{Analysis}
We take a look at 2 examples in this section to perform inference on the generated natural language summaries of the methods. The examples are handpicked for illustration purposes only. 
\begin{enumerate}
    \item 
    Original Code: 
\begin{lstlisting}
public PartVO getChild(){
        return child;
    }
\end{lstlisting}
Tokenized input to the model: "public part vo get child return child".

Reference comment: "get the child part".

deepSumm prediction: "gets the child".

\begin{figure}[htp]
    \includegraphics[width=10cm]{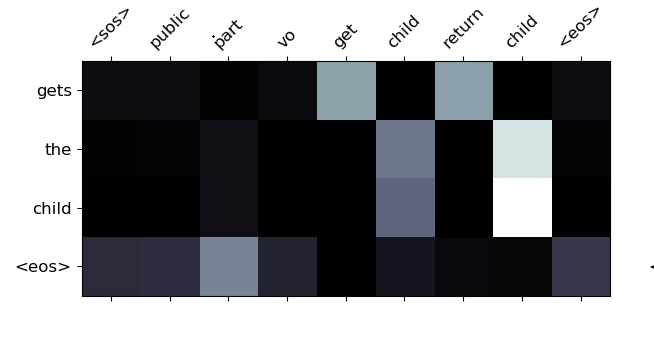}
    \caption{Attention distribution for the first example}
    \label{fig:attn1}
\end{figure}
We see this example as an acceptable generation of the summary as compared to the reference. The weights on the attention distribution shows that to generate "gets" the decoder is paying attention to "get" and "return" tokens from the source sequence which is justifiable. It follows the suit for the predicting "child". 

    \item 
    Original Code: 
\begin{lstlisting}
public double getOxygenConsumptionRate() {
        return getValueAsDouble(OXYGEN_CONSUMPTION_RATE);
    }
        
\end{lstlisting}
Tokenized input to the model: "public double get oxygen consumption rate return get value as double oxygen consumption rate".

Reference comment: "gets the oxygen consumption rate".

deepSumm prediction: "returns the water consumption rate for this object".

\begin{figure}[htp]
    \includegraphics[width=10cm]{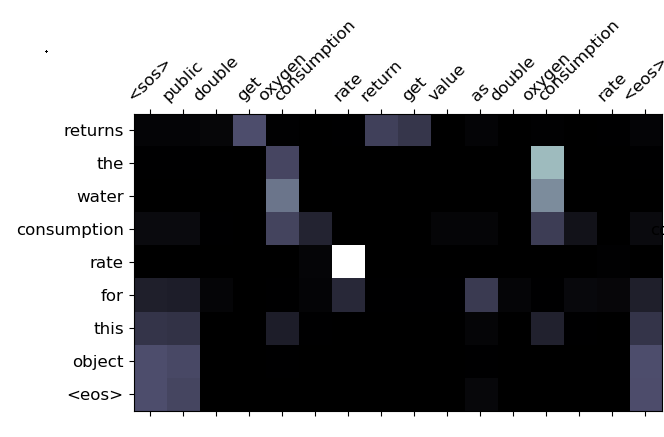}
    \caption{Attention distribution for the 2nd example}
    \label{fig:attn2}
\end{figure}
We see this example as a problematic generation of the summary as compared to the reference. Looking at the weights, the model while decoding the third token, paid attention to the consumption and predicted water. We think that the greedy algorithm paid the price of the faulty search here. 
   \item 
    Original Code: 
\begin{lstlisting}
public void setMaximumColorDepth(String value) {
        this.maximumColorDepth = value;
    }        
\end{lstlisting}
Tokenized input to the model: "public double get oxygen consumption rate return get value as double oxygen consumption rate".

Reference comment: "gets the oxygen consumption rate".

deepSumm prediction: "returns the water consumption rate for this object".

Recurrent attention(Small model) prediction: "sets the maximum color depth of the color"
\begin{figure}[htp]
    \includegraphics[width=10cm]{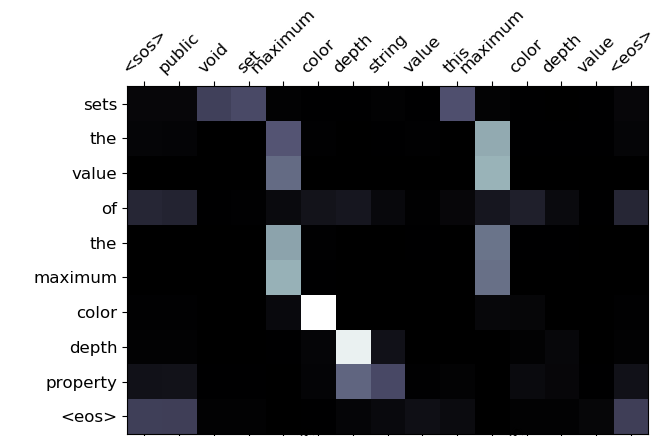}
    \caption{Attention distribution for the 3rd example}
    \label{fig:attn3}
\end{figure}
We see this example as a good generation of the summary as compared to the reference and as compared tot he attentive RNN model which had a poor prediction.

\end{enumerate}

\section{Conclusion and Future work}

With this work, we presented a model that effectively generates natural language summaries from program methods written in Java. Furthermore, we were able to create two contrasting models, one that uses Recurrence and other that uses state of the art, Transformer architecture. We observed the benefits of Transformer as cited by Vaswani \textit{et al.} \cite{vaswani2017attention} first hand by running various experiments on both these models. Our transformer model was able to outperform recurrence based models by more that a BLEU score in the same dataset. The training time also improved. Empirically with this dataset by a factor of half when compared with recurrence. While we were not able to achieve state of the art results in the task but we think that employing effective structural information along with programming constructs in representing the methods will come handy along with more efficient strategy in exploring the gradient space and training the network for longer time. Furthermore, exploring ensemble methods in NLP as done by LeClair \textit{et al.} \cite{alex2019neural} will also be a great strategy. We think, that these will be helpful next steps to take the next step in this line of work. 
 
\bibliographystyle{unsrtnat}
\bibliography{deepsumm}

\begin{thebibliography}{17}
\providecommand{\natexlab}[1]{#1}
\providecommand{\url}[1]{\texttt{#1}}
\expandafter\ifx\csname urlstyle\endcsname\relax
  \providecommand{\doi}[1]{doi: #1}\else
  \providecommand{\doi}{doi: \begingroup \urlstyle{rm}\Url}\fi

\bibitem[Forward and Lethbridge(2002)]{forward}
Andrew Forward and Timothy~C. Lethbridge.
\newblock The relevance of software documentation, tools and technologies: A
  survey.
\newblock In \emph{Proceedings of the 2002 ACM Symposium on Document
  Engineering}, DocEng ’02, page 26–33, New York, NY, USA, 2002.
  Association for Computing Machinery.
\newblock ISBN 1581135947.
\newblock \doi{10.1145/585058.585065}.
\newblock URL \url{https://doi.org/10.1145/585058.585065}.

\bibitem[Haiduc et~al.(2010)Haiduc, Aponte, Moreno, and Marcus]{HaiducAMM10}
Sonia Haiduc, Jairo Aponte, Laura Moreno, and Andrian Marcus.
\newblock On the use of automated text summarization techniques for summarizing
  source code.
\newblock In Giuliano Antoniol, Martin Pinzger, and Elliot~J. Chikofsky,
  editors, \emph{WCRE}, pages 35--44. IEEE Computer Society, 2010.
\newblock ISBN 978-0-7695-4123-5.
\newblock URL
  \url{http://dblp.uni-trier.de/db/conf/wcre/wcre2010.html#HaiducAMM10}.

\bibitem[Moreno et~al.(2013)Moreno, Aponte, Sridhara, Marcus, Pollock, and
  Vijay-Shanker]{Moreno2013AutomaticGO}
Laura Moreno, Jairo Aponte, Giriprasad Sridhara, Andrian Marcus, Lori~L.
  Pollock, and K.~Vijay-Shanker.
\newblock Automatic generation of natural language summaries for java classes.
\newblock \emph{2013 21st International Conference on Program Comprehension
  (ICPC)}, pages 23--32, 2013.

\bibitem[Iyer et~al.(2016)Iyer, Konstas, Cheung, and
  Zettlemoyer]{iyer-etal-2016-summarizing}
Srinivasan Iyer, Ioannis Konstas, Alvin Cheung, and Luke Zettlemoyer.
\newblock Summarizing source code using a neural attention model.
\newblock In \emph{Proceedings of the 54th Annual Meeting of the Association
  for Computational Linguistics (Volume 1: Long Papers)}, pages 2073--2083,
  Berlin, Germany, August 2016. Association for Computational Linguistics.
\newblock \doi{10.18653/v1/P16-1195}.
\newblock URL \url{https://www.aclweb.org/anthology/P16-1195}.

\bibitem[Angeli et~al.(2010)Angeli, Liang, and Klein]{angeli-etal-2010-simple}
Gabor Angeli, Percy Liang, and Dan Klein.
\newblock A simple domain-independent probabilistic approach to generation.
\newblock In \emph{Proceedings of the 2010 Conference on Empirical Methods in
  Natural Language Processing}, pages 502--512, Cambridge, MA, October 2010.
  Association for Computational Linguistics.
\newblock URL \url{https://www.aclweb.org/anthology/D10-1049}.

\bibitem[Konstas and Lapata(2013)]{Ioannis}
Ioannis Konstas and Mirella Lapata.
\newblock A global model for concept-to-text generation.
\newblock \emph{J. Artif. Int. Res.}, 48\penalty0 (1):\penalty0 305–346,
  October 2013.
\newblock ISSN 1076-9757.

\bibitem[Bahdanau et~al.(2014)Bahdanau, Cho, and Bengio]{bahdanau2014neural}
Dzmitry Bahdanau, Kyunghyun Cho, and Yoshua Bengio.
\newblock Neural machine translation by jointly learning to align and
  translate, 2014.

\bibitem[Wei et~al.(2019)Wei, Li, Xia, Fu, and Jin]{wei2019code}
Bolin Wei, Ge~Li, Xin Xia, Zhiyi Fu, and Zhi Jin.
\newblock Code generation as a dual task of code summarization, 2019.

\bibitem[Hu et~al.(2018)Hu, Li, Xia, Lo, and Jin]{Hu}
Xing Hu, Ge~Li, Xin Xia, David Lo, and Zhi Jin.
\newblock Deep code comment generation.
\newblock In \emph{Proceedings of the 26th Conference on Program
  Comprehension}, ICPC ’18, page 200–210, New York, NY, USA, 2018.
  Association for Computing Machinery.
\newblock ISBN 9781450357142.
\newblock \doi{10.1145/3196321.3196334}.
\newblock URL \url{https://doi.org/10.1145/3196321.3196334}.

\bibitem[LeClair et~al.(2019)LeClair, Jiang, and McMillan]{alex2019neural}
Alexander LeClair, Siyuan Jiang, and Collin McMillan.
\newblock A neural model for generating natural language summaries of program
  subroutines, 2019.

\bibitem[LeClair and McMillan(2019)]{alex2019recommendations}
Alexander LeClair and Collin McMillan.
\newblock Recommendations for datasets for source code summarization, 2019.

\bibitem[Luong et~al.(2015)Luong, Pham, and Manning]{luong2015effective}
Minh-Thang Luong, Hieu Pham, and Christopher~D. Manning.
\newblock Effective approaches to attention-based neural machine translation,
  2015.

\bibitem[Sutskever et~al.(2014)Sutskever, Vinyals, and
  Le]{sutskever2014sequence}
Ilya Sutskever, Oriol Vinyals, and Quoc~V. Le.
\newblock Sequence to sequence learning with neural networks, 2014.

\bibitem[Vaswani et~al.(2017)Vaswani, Shazeer, Parmar, Uszkoreit, Jones, Gomez,
  Kaiser, and Polosukhin]{vaswani2017attention}
Ashish Vaswani, Noam Shazeer, Niki Parmar, Jakob Uszkoreit, Llion Jones,
  Aidan~N. Gomez, Lukasz Kaiser, and Illia Polosukhin.
\newblock Attention is all you need, 2017.

\bibitem[Ba et~al.(2016)Ba, Kiros, and Hinton]{ba2016layer}
Jimmy~Lei Ba, Jamie~Ryan Kiros, and Geoffrey~E. Hinton.
\newblock Layer normalization, 2016.

\bibitem[Papineni et~al.(2002)Papineni, Roukos, Ward, and Zhu]{Bleu}
Kishore Papineni, Salim Roukos, Todd Ward, and Wei-Jing Zhu.
\newblock Bleu: A method for automatic evaluation of machine translation.
\newblock In \emph{Proceedings of the 40th Annual Meeting on Association for
  Computational Linguistics}, ACL ’02, page 311–318, USA, 2002. Association
  for Computational Linguistics.
\newblock \doi{10.3115/1073083.1073135}.
\newblock URL \url{https://doi.org/10.3115/1073083.1073135}.

\bibitem[Contributors.(2018)]{bleu_score}
Torch Contributors.
\newblock Torchtext.data.metrics, 2018.
\newblock URL \url{https://pytorch.org/text/data_metrics.html#bleu-score}.

\end{thebibliography}

\end{document}